\documentclass{article} % For LaTeX2e
\usepackage{nips14submit_e,times}
\usepackage{url}

\usepackage{amssymb}
\usepackage{amsmath}
\usepackage{graphicx}
\usepackage{algorithm}
\usepackage{caption}
\usepackage{subcaption}
\usepackage{algpseudocode}
\usepackage{color}

\title{SAME but Different: Fast and High-Quality Gibbs Parameter Estimation}

\author{
Huasha Zhao \\
Computer Science Division\\
UC Berkeley\\
Berkeley, CA 94720 \\
\texttt{hzhao@cs.berkeley.edu} \\
\And
Biye Jiang \\
Computer Science Division\\
UC Berkeley\\
Berkeley, CA 94720 \\
\texttt{bjiang@cs.berkeley.edu} \\
\AND
John Canny \\
Computer Science Division\\
UC Berkeley\\
Berkeley, CA 94720 \\
\texttt{jfc@cs.berkeley.edu} \\
}

\nipsfinalcopy % Uncomment for camera-ready version

\newcommand{\qed}{\nobreak \ifvmode \relax \else
      \ifdim\lastskip<1.5em \hskip-\lastskip
      \hskip1.5em plus0em minus0.5em \fi \nobreak
      \vrule height0.75em width0.5em depth0.25em\fi}

\begin{document}

\maketitle

\begin{abstract}
Gibbs sampling is a workhorse for Bayesian inference but has several
limitations when used for parameter estimation, and is often much
slower than non-sampling inference methods. SAME (State Augmentation
for Marginal Estimation) \cite{Doucet99,Doucet02} is an approach to MAP
parameter estimation which gives improved parameter estimates over
direct Gibbs sampling. SAME can be viewed as cooling the posterior
parameter distribution and allows annealed search for the MAP
parameters, often yielding very high quality (lower loss) estimates. But it does
so at the expense of additional samples per iteration and generally
slower performance. On the other hand, SAME dramatically increases the
parallelism in the sampling schedule, and is an excellent match for
modern (SIMD) hardware. In this paper we explore the application of
SAME to graphical model inference on modern hardware. We show that
combining SAME with factored sample representation (or approximation)
gives throughput competitive with the fastest symbolic methods, but
with potentially better quality. We describe experiments on Latent
Dirichlet Allocation, achieving speeds similar to the fastest
reported methods (online Variational Bayes) and lower cross-validated
loss than other LDA implementations. The method is simple to implement and
should be applicable to many other models. 

\end{abstract}

\section{Introduction}

Many machine learning problems can be formulated as inference
on a joint distribution $P(X,Z,\Theta)$ where $X$ represents observed
data, $\Theta$ a set of parameters, and $Z$ represents latent variables. 
Both $\Theta$ and $Z$ are latent in general, but it is useful to distinguish
between them - $\Theta$ encodes which of a class of models represents
the current situation, while $Z$ represent local labels or missing data.
One generally wants to optimize over $\Theta$ while marginalizing over $Z$.
And the output of the algorithm is a value or distribution over $\Theta$
while the $Z$ are often ignored. 

Gibbs sampling is a very general approach to posterior estimation for
$P(X,Z,\Theta)$, but it provides samples only rather than MAP
estimates.  But therein lies a problem: sampling is a sensible
approach to marginal estimation, but can be a very inefficient approach to
optimization. This is particularly true when the dimension of $\Theta$
is large compared to $X$ (which is true e.g. in Latent Dirichlet
Allocation and probabilistic recommendation algorithms).  Such models
have been observed to require many samples (thousands to hundreds of
thousands) to provide good parameter estimates. Hybrid approaches such
as Monte-Carlo EM have been developed to address this issue - a
Monte-Carlo method such as Gibbs sampling is used to estimate the
expected values in the E-step while an optimization method is applied
to the parameters in the M-step. But this requires a separate
optimization strategy (usually gradient-based), a way to compute the
dependence on the parameters symbolically, and analysis of the
accuracy of the E-step estimates.

SAME (State Augmentation for Marginal Estimation)
\cite{Doucet99,Doucet02} is a simple approach to MAP parameter
estimation that remains within the Gibbs framework\footnote{SAME is a
  general approach to MCMC MAP estimation, but in this paper we will
  focus on its realization in Gibbs samplers}. SAME replicates the
latent state $Z$ with additional states. This has the effect of
``cooling'' the marginal distribution on $\Theta$, which sharpens its
peaks and causes $\Theta$ samples to approach local optima.  The
conditional distribution $P(Z|X,\Theta)$ remains the same, so we are
still marginalizing over a full distribution on $Z$.  By
making the temperature a controllable parameter, the 
parameter estimates can be annealed to reach better local optima.
In both \cite{Doucet99,Doucet02} and the present paper we find that
this approach gives better estimates than competing approaches. 
The novelty of the present paper is showing that SAME estimation
can be {\em very fast}, and competitive with the fastest symbolic
meethods. Thus it holds the potential to be the method of choice
for many inference problems.

Specifically, we define a new joint distribution
\begin{equation}
P^{\prime}(X,\Theta,Z^{(1)},\ldots,Z^{(m)})=\prod_{j=1}^{m} P(X,\Theta,Z^{(j)})
\end{equation}
which models $m$ copies of the original system with tied parameters
$\Theta$ and independent latent variable blocks $Z^{(1)},\ldots,Z^{(m)}$. The marginalized
conditional $P^{\prime}(\Theta|X) = P^{\prime}(X,\Theta)/P(X)$. And
\begin{equation}
P^{\prime}(X,\Theta) = \int_{Z^{(1)}}\cdots\int_{Z^{(m)}}\prod_{j=1}^m P(X,\Theta,Z^{(j)})~ dZ^{(1)}\cdots dZ^{(m)}
=\prod_{j=1}^m P(X,\Theta) = P^m(X,\Theta)
\end{equation}
where $P(X,\Theta)=\int_Z P(X,\Theta,Z)~dZ$.  So $P^{\prime}(\Theta|X)
= P^m(X,\Theta)/P(X)$ which is up to a constant factor equal to
$P^m(\Theta|X)$, a power of the original marginal parameter distribution. 
Thus it has the same optima, including the global optimum, but its
peaks are considerably sharpened. In what follows we will often demote $X$
to a subscript since it fixed, writing $P(\Theta|Z,X)$ as $P_X(\Theta|Z)$ etc.

This new distribution can be written as a Gibbs distribution on $\Theta$, as
$P^m(\Theta,X) = \exp(-mg_X(\Theta)) = \exp(-g_X(\Theta)/(kT))$, from
which we see that $m=1/(kT)$ is an inverse temperature parameter ($k$
is Boltzmann's constant). Increasing $m$ amounts to cooling the
distribution.

Gibbs sampling from the new distribution is usually straightforward
given a sampler for the original. It is perhaps not obvious why
sampling from a more complex system could improve performance, but
we have added considerable parallelism since we can sample various
``copies'' of the system concurrently. It will turn out this approach
is complementary to using a factored form for the posterior.
Together these methods gives us orders-of-magnitude speedup over other
samplers for LDA.

The rest of the paper is organized as follows. Section 2 summarizes
related work on parameter inference for probabilistic models and their
limitations. Section 3 introduces the SAME sampler.  We discuss in
Section 4 a factored approximation that considerably accelerates
sampling.  A hardware-optimized implementation of the algorithm for
LDA is described in Section 5. Section 6 presents the experimental
results and finally Section 7 concludes the paper.

\section{Related Work}

\subsection{EM and related Algorithms}

The Expectation-Maximization (EM) algorithm \cite{dempster1977maximum}
is a popular method for parameter estimation of graphical models of
the form we are interested in. The EM algorithm alternates between
updates in the expectation (E) step and maximization (M) step. The
E-step marginalizes out the latent variables and computes the
expectation of the likelihood as a function of the parameters.  The E
step computes a $Q$ function $Q(\Theta^\prime | \Theta) =
E_{Z|\Theta}(\log P_X(Z, \Theta^\prime))$ to be optimized in the
M-step.

For EM to work, one has to compute the expectation of the sufficient
statistics of the likelihood function, where the expectation is over
$Z|\Theta$. It also requires a method to optimize the $Q$ function. In
practice, the iterative update equations can be hard to derive.
Moreover, the EM algorithm is a gradient-based method, and therefore
is only able to find locally-optimal solutions.
% empirical results suggest that EM cannot always produces as accurate solutions as sampling approaches. 

Variational Bayes (VB) \cite{jordan1999introduction} is an EM-like
algorithm that uses a parametric approximation to the posterior
distribution of both parameters and other latent variables, and
attempts to optimize the fit (e.g. using KL-divergence) to the
observed data.  Parameter estimates are usually taken from the means
of the parameter approximations (the hyperparameters), unlike EM where
point parameter estimates are used.  It is common to assume a
coordinate-factored form for this approximate posterior.  The factored
form simplifies inference, but makes strong assumptions about the
distribution (effectively eliminating interactions). It makes most
sense for parameter estimates, but is a strong constraint to apply to
other latent variables.  VB has been criticized for lack of quality
(higher loss) on certain problems because of this.  Nevertheless it
works well for many problems (e.g. on LDA).  This motivated us to
introduce a similar factored approximation in our method described
laster.

\subsection{Gibbs Sampling}

Gibbs samplers \cite{geman1984stochastic} work by stepping through
individual (or blocks of) latent variables, and in effect perform a
simulation of the stochastic system described by the joint
distribution. The method is simple to apply on graphical models since
each variable is conditioned locally by its Markov blanket.  They are
easy to define for conjugate distributions and can generalize to
non-conjugate graphical models using e.g. slice sampling.  Gibbs
samplers are often the first method to be derived, and sometimes the
only practical learning strategy for complex graphical models.

Gibbs samplers only require that the joint density is strictly positive
over the sample space which is known as the Gibbs
distribution. However, Gibbs sampling only gives samples from the
distribution of $\Theta$, it can be difficult to find ML or MAP values
from those samples. Furthermore, it can be very slow, especially for
models with high dimensions. Our results suggest this slow convergence
is often due to large variance in the parameters in standard samplers. 

%% Blocking \cite{liu1994collapsed} and collapsing
%% \cite{liu1994collapsed} are two common techniques to accelerate the
%% convergence. In this article, we adapt the block sampling for
%% simulating the hidden variable $Z$ when possible. We also collapse
%% $\Theta$ for our LDA implementation.

\subsection{Monte Carlo EM}
Monte Carlo EM \cite{wei1990monte} is a hybrid approach that uses
MCMC (e.g. Gibbs sampling) to approximate the expected
value $E_{Z|\Theta}(\log P_X(Z, \Theta^\prime))$ with the mean of the
log-likelihood of the samples.
%\begin{align}
%Q(\Theta^\prime | \Theta) = \sum\limits_{i=1}^{m}\log P_X(Z_{i}, \Theta^\prime),
%\end{align}
%where $Z_{(i)}$'s are drawn from $Z|\Theta$ independently.
The method has to optimize $Q(\Theta^\prime | \Theta)$ using a
numerical method (conjugate gradient etc.). Like standard EM, it can suffer from convergence problems, and may only find
a local optimum of likelihood.

\subsection{Message-Passing Methods}
Belief propagation \cite{YW2000} and Expectation propagation
\cite{TM2001} use local (node-wise) updates to infer posterior
parameters in graphical models in a manner reminiscent of Gibbs
sampling. But they are exact only for a limited class of models.
Recently variational message-passing \cite{WB2005} has extended the
class of models for which parametric inference is possible to
conjugate-exponential family graphical models. However similar to
standard VB, the method uses a coordinate-factored approximation to
the posterior which effectively eliminates interactions (although they
can be added at high computational cost by using a factor graph). It
also finds only local optima of the posterior parameters.

\section{SAME Parameter Estimation}

SAME estimation involves sampling multiple $Z$'s independently and
inferring $\Theta$ using the aggregate of $Z$'s.

\subsection{Method}
We use the notation $Z_{-i} =
Z_1,\ldots,Z_{i-1},Z_{i+1},\ldots,Z_n$ and similarly for
$\Theta_{-i}$.

%For many models, each coordinate of $Z$ is independent from each other given $\Theta$ (the graph is two color-able), so sampling conditioned $Z$ can be highly parallel.

\begin{algorithm}
\caption{Standard Gibbs Parameter Estimation}
\begin{algorithmic}[1]
\State initialize parameters $\Theta$ randomly, then in some order:
\State{Sample $Z_i \sim P_X(Z_i | Z_{-i}, \Theta)$} 
\State{Sample $\Theta_i \sim P_X(\Theta_i | \Theta_{-i}, Z)$} 
\end{algorithmic}
\label{SGS}
\end{algorithm}

%% If we repeatedly sample $Z$ for $m$ times from its conditionals and in
%% turn sample $\Theta$ from multiple $Z$'s using Bayesian inference, we
%% progressively reduce variance and effectively preserve more
%% information, and in the limit approach the conditional mean. This
%% algorithm is described in Algorithm \ref{MSGS}. For large $m$,
%% conditional mean and mode are very close. Our method effectively finds
%% the MAP estimation of the parameter.

%\begin{figure*}[t!]
%\centering
%\includegraphics[scale = 0.6]{framework.png}
%\caption{Standard Gibbs (top) vs. Cooled Gibbs (bottom)}
%\label{framework}
%\end{figure*}

\begin{algorithm}
\caption{SAME Parameter Estimation}
\begin{algorithmic}[1]
\State initialize parameters $\Theta$ randomly, and in some order:
\State{Sample $Z^{(j)}_{i} \sim P_X(Z_i^{(j)} | Z^{(j)}_{-i}, \Theta)$} 
\State{Sample $\Theta_i \sim P_X(\Theta_i | \Theta_{-i}, Z^{(1)}, \ldots, Z^{(m)})$} 
\end{algorithmic}
\label{MSGS}
\end{algorithm}

Sampling $Z_i^{(j)}$ in the SAME sampler is exactly the same as for
the standard sampler. Since the groups $Z^{(j)}$ are independent of
each other, we can use the sampling function for the original
distribution, conditioned only on the other components of the same
group: $Z_{-i}^{(j)}$.

Sampling $\Theta_i$ is only slightly more complicated.  We want to sample from
\begin{equation}
P_X(\Theta_i | \Theta_{-i}, Z^{(1)}, \ldots, Z^{(m)}) = P_X(\Theta, \overline{Z})/P_X(\Theta_{-i},\overline{Z}) 
\end{equation}
where $\overline{Z}=Z^{(1)},\ldots,Z^{(m)}$ and if we ignore the normalizing constants:
\begin{equation}
P_X(\Theta, \overline{Z})/P_X(\Theta_{-i},\overline{Z}) \propto P_X(\Theta, \overline{Z}) = \prod_{j=1}^m P_X(\Theta,Z^{(j)}) \propto \prod_{j=1}^m P_X(\Theta_i|\Theta_{-i},Z^{(j)})
\end{equation}
which is now expressed as a product of conditionals from the original
sampler $P_X(\Theta_i|\Theta_{-i},Z^{(j)})$. Inference in the new
model will be tractable if we are able to sample from a product of the
distributions $P_X(\Theta_i|\Theta_{-i},Z^{(j)})$. This will be true
for many distributions. e.g. for exponential family distributions in
canonical form, the product is still an exponential family member. A
product of Dirichlet distributions is Dirichlet etc., and in general
this distribution represents the parameter estimate obtained by
combining evidence from independent observations. The normalizing constant
will usually be implied from the closed-form parameters of this distribution. 

Adjusting sample number $m$ at different iterations allows annealing of the 
estimate. 

\section{Coordinate-Factored Approximation}
Certain distributions (including LDA) have the property that the
latent variables $Z_i$ are independent given $X$ and $\Theta$. That is
$P(Z_i|Z_{-i},X,\Theta) = P(Z_i|X,\Theta)$. Therefore the
$Z_i$'s can be sampled (without approximation) in parallel. Furthermore, rather than a single
sample from $P(Z_i|X,\Theta)$ (e.g. a categorical sample for a discrete 
$Z_i$) we can construct a SAME Gibbs sampler by taking $m$ samples. 
These samples will now have a multinomial distribution with count $m$ and
probability vector $P(Z_i|X,\Theta)$. 
Let $\hat{Z}_i(v)$ denote the count for $Z_i = v$ among the $m$ samples, and
$P(Z_i=v|X,\Theta)$ denote the conditional probability that $Z_i = v$. 

We can introduce still more parallelism by randomizing the order in
which we choose which $Z_i$ from which to sample.  The count $m$ for
variable $Z_i$ is then replaced by random variable $\hat{m}\sim \text{Poisson}(m)$
and the coordinate-wise distributions of $\hat{Z}_i$ become independent
Poisson variables:

\begin{equation}
\hat{Z}_i(v) \sim \text{Poisson}(mP(Z_i=v|X,\Theta))
\end{equation}

when the $Z_i$ are independent given $X,\Theta$, the counts
$\hat{Z}_i(v)$ fully capture the results of taking the $m$
(independent) samples.  These samples can be generated very fast, and
completely in parallel. $m$ is no longer constrained to be an integer,
or even to be $>1$. Indeed, each sample no longer corresponds to
execution of a block of code, but is simply an increment in the value
$m$ of the Poisson random number generator. In LDA, it is almost as
fast to generate all $m$ samples for a single word for a large $m$ (up
to about 100) as it is to generate one sample. This is a source of
considerable speedup in our LDA implementation. 

For distributions where the $Z_i$ are not independent, i.e. when
$P(Z_i|Z_{-i},X,\Theta) \neq P(Z_i|X,\Theta)$, we can still perform
independent sampling {\bf as an approximation}.  This approach is
quite similar to the coordinate-factored approximation often used in
Variational Bayes. We leave the details to a forthcoming paper.

\section{Implementation of SAME Gibbs LDA}

%\subsection{Parallel Sampling under SIMD Architecture}
%Random Number Generator (RNG) is key primitive for simulating large Bayesian networks using Monte Carlo method such as Gibbs sampling. It is a critical component that determines the performance of such simulations. The Single Instruction Multiple Data (SIMD) architecture for RNG is a natural fit for the parallel sampling paradigm to accelerate simulation. SIMD exists on both CPUs and GPUs. However, GPU has much higher levels of parallelism and significant memory bandwidth. GPUs have seen much service in scientific computing, but several factors have hampered their use as general computing accelerators, especially in machine learning. Primarily these were slow transfers from CPU to GPU memory, and limited GPU memory size. However, transfer speeds have improved significantly thanks to PCI-bus improvements, and memory capacity now is quite good (2-4 GBytes typical).
%
%\begin{figure}
%\centering
%\includegraphics[scale = 0.5]{rand.png}
%\caption{Performance of RNG on GPU and CPU}
%\label{rand}
%\end{figure} 
%
%Figure \ref{rand} illustrates the benchmark of RNG on two SIMD supported devices: an Intel E5-2660 CPU and an Nvidia GTX-690 GPU. Only one core of each device is used for the benchmark. As we can see, both devices demonstrate considerable parallelism, they generate up to $10^4$ and $10^6$ random number at a fixed cost for CPU and GPU respectively. We also noticed that GPU is 20x faster than CPU for generating sufficient large number of random numbers. 
Our SAME LDA sampler implementation is described in Algorithm \ref{LDA}.
Samples are taken directly from Poisson distributions (line $7$ of the algorithm) as described
earlier. 

%In the following, we first review LDA and give the collapsed
%sampling update equation (Equation \ref{update}) for cooled Gibbs, and
%then present several techniques to remove potential bottlenecks of its
%scalable implementation.

\subsection{Latent Dirichlet Allocation}
\newcommand*{\LargerCdot}{\raisebox{-0.25ex}{\scalebox{1.2}{$\cdot$}}}
%
%\begin{align}
%P(X,Z,\theta, \phi | \alpha, \beta)
%&=\prod_{d=1}^{D} \prod_{k=1}^{K}{\theta_{d,k}^{\alpha}} \prod_{i=1}^{N_d}{\theta_{d, z_{d,i}}} \prod_{k=1}^{K} \prod_{w=1}^{W} {\phi_{k,w}^{\beta}} \prod_{d=1}^{D}\prod_{i=1}^{N_d}{\phi_{z_{d,i}, y_{d,i}}} \\\nonumber
%&=\prod_{d=1}^{D} \prod_{k=1}^{K}{\theta_{d,k}^{\alpha}} \prod_{k=1}^{K}{\theta_{d, k}^{c_{k, d,\LargerCdot }}} \prod_{k=1}^{K} \prod_{w=1}^{W} {\phi_{k,w}^{\beta}} \prod_{w=1}^{W} {\phi_{k,j}^{c_{k,\LargerCdot,w}}} \\\nonumber
%&=\prod_{d=1}^{D} \prod_{k=1}^{K}{\theta_{d,k}^{\alpha + c_{k, d,\LargerCdot }}} \prod_{k=1}^{K}\prod_{w=1}^{W}{\phi_{k,j}^{\beta + c_{k,\LargerCdot,w}}}.
%\end{align}

%$c$ are count variables determined by $X$ and $Z$, and are defined as, 
%\begin{align}
%c_{k,d,w} = \sum_{i=1}^{N_d} 1(z_{d, i}=k \text{ and } x_{d, i}=w), & \\\nonumber
%c_{\LargerCdot, d, w} = \sum_{k=1}^{K} c_{k,d,w},c_{k, \LargerCdot,  w} = \sum_{d=1}^{D} c_{k,d,w} \text{ and } c_{k, d, \LargerCdot} =& \sum_{w=1}^{W} c_{k,d,w}.
%\end{align}

LDA is a generative process for modeling a collection of documents in
a corpus. In the LDA model, the words $X=\{x_{d,i}\}$ are observed,
the topics $Z=\{z_{d,i}\}$ are latent variables and parameters are
$\Theta=(\theta, \phi)$ where $\theta$ is document topic distributions
and $\phi$ is word topic distributions. Subscript $d, i$ denotes document $d$ and its $i^{th}$ word. Given $X$ and $\Theta$,
$z_{d,i}$ are independent, which also implies that we can sample from
$Z$ without any information about $Z$ from previous samples. A similar result
holds for the parameters, which can be sampled given only the current counts
from the $Z_i$. We can therefore alternate parameter and latent variable inference using a pair of
samplers:
$P_X(Z|\theta, \phi)$ and $P_X(\theta, \phi|Z)$. Such blocked sampling maximizes
parallelism and works very well with modern SIMD hardware. 

%The conditional distribution of $Z$ is proportional to the likelihood given fixed variables,
%\begin{align}
%P(Z|X, \theta, \phi, \alpha, \beta)&=\frac{P(Z, X, \theta, \phi | \alpha, \beta)}{P(X, \theta, \phi | \alpha, \beta)}
%\propto P(Z, X, \theta, \phi | \alpha, \beta).
%\end{align}

%Conditional $Z_{d, i}$'s are independent. After dropping fixed terms, $Z_{d, i}$ follows a discrete distribution,
%\begin{align}
%P_X(z_{d,i}| \theta, \phi, \alpha, \beta) \propto \prod_{k=1}^{K} \theta_{d,k}^{1(z_{d,i}=k)} \prod_{k=1}^{K}\phi_{k,x_{d,i}}^{1(z_{d,i}=k)}.
%\end{align}
% 
%Similarly, the conditional distribution of $\Theta$ is proportional to the likelihood,  
%\begin{align}
%P(\theta, \phi|X, Z, \alpha, \beta)&=\frac{P(Z, X, \theta, \phi | \alpha, \beta)}{P(X, Z | \alpha, \beta)}
%\propto P(Z, X, \theta, \phi | \alpha, \beta)
%\end{align}
%
%Conditional $\Theta$ follows a Dirichlet distribution, 
%
%\begin{align}
%P(\theta, \phi|X, Z, \alpha, \beta)
%\propto \prod_{d=1}^{D} \prod_{k=1}^{K}{\theta_{d,k}^{\alpha + c_{k, d,\LargerCdot }}} \prod_{k=1}^{K}\prod_{w=1}^{W}{\phi_{k,j}^{\beta + c_{k,\LargerCdot,w}}}.
%\end{align}

The sampling of $z_{d,i}$'s uses the Poisson formula derived earlier.  The
conditional distributions $P_X(\theta, \phi|z_{d,i})$ are multiple
independent Dirichlet's. In practice, we collapse out $(\theta,\phi)$,
so we in effect sample a new $z_{d,i}^t$ given the $z^{t-1}$ from a
previous iteration.  The
update sampler follows a Dirichlet compound multinomial distribution
(DCM) and can be derived as,
\begin{align}
P_X(z^t_{d,i} | z^{t-1}, \alpha, \beta) 
&=\frac{{c_{k,d,\LargerCdot}}/{m}+\alpha}{c_{\LargerCdot,d,\LargerCdot}/m+K\alpha} \frac{c_{k,\LargerCdot, w}/m+\beta}{c_{k, \LargerCdot,\LargerCdot}/m+W\beta},
\label{update}
\end{align}
where $W$ and $K$ are numbers of words and topics respectively, $c$ are counts across all samples and all documents which are defined as,
\begin{align}
c_{k,d,w} = \sum_{j=1}^{m}\sum_{i=1}^{N_d} 1(z^{(j)}_{d, i}=k \text{ and } x_{d, i}=w),
%c_{\LargerCdot, d, w} = \sum_{k=1}^{K} c_{k,d,w},c_{k, \LargerCdot,  w} = \sum_{d=1}^{D} c_{k,d,w} \text{ and } c_{k, d, \LargerCdot} = \sum_{w=1}^{W} c_{k,d,w}.
\end{align}
where $N_d$ is the number of documents and superscript $(j)$ of $z$ denotes the $j^{th}$ sample of the hidden topic variable. In Equation \ref{update}, dot(s) in the subscript of $c$ denotes integrating (summing) over that dimension(s), for example, $ c_{k, d, \LargerCdot} = \sum_{w=1}^{W} c_{k,d,w}$. As shown in Equation \ref{update}, sample counts are sufficient statistics for the update formula.

\subsection{Mini-batch Processing}

For scalability (to process datasets that will not fit in memory), we
implement LDA using a mini-batch update strategy. In mini-batch
processing, data is read from a dataset and processed in
blocks. Mini-batch algorithms have been shown to be very efficient for
approximate inference \cite{bottou2007tradeoffs, bottou2010large}. 

In Algorithm \ref{LDA}, $D_t$ is a sparse (nwords x ndocs) matrix that
encodes the subset of documents (mini-batch) to process at period
$t$. The global model (across mini-batches) is the word-topic matrix
$\phi$. It is updated as the weighted average of the current model and
the new update, see line $14$. The weight is determined by $\rho_t =
(\tau_0 + t)^{-\gamma}$ according to \cite{hoffman2010online}. We do not
explicitly denote passes over the dataset. The data are treated instead
as an infinite stream and we examine the cross-validated likelihood
as a function of the number of mini-batch steps, up to $t_{max}$.

\begin{algorithm}
\caption{SAME Gibbs LDA}
\begin{algorithmic}[1]
\For{$t=0 \to t_{max}$} 
\State $\hat \theta=0$; $\hat \phi=0$; $\rho_t = (\tau_0 + t)^{-\gamma}$
\State $\mu = SDDMM(\theta, \phi, D_t)$
\ForAll{document-word pair $(d, w)$ in mini-batch $D_t$ $\mathbf{parallel}$}
\For{$k=1 \to K$ $\mathbf{parallel}$ } 
\State{$\lambda = \theta_{d,k} \phi_{k, w} / \mu_{d, w}$}
\State{sample $z \sim \text{Poisson}(\lambda \cdot m)$} 
\State $\hat\theta_{d,k} = \hat\theta_{d,k} + z/m$
\State $\hat\phi_{k,w} = \hat\phi_{k,w} + z/m$
\EndFor
\EndFor
\State $\theta = \hat \theta + \alpha$; $\phi = \hat \phi + \beta$ \State normalize $\hat \phi$ along the word dimension
\State{$\phi = (1-\rho_t)\phi + \rho_t\hat\phi$}
\EndFor
\end{algorithmic}
\label{LDA}
\end{algorithm}

\subsection{GPU optimizations}
GPUs are extremely well-suited to Gibbs sampling by virtue of their
large degree of parallelism, and also because of their extremely high
throughput in non-uniform random number generation (thanks to
hardware-accelerated transcendental function evaluation).  For best
performance we must identify and accelerate the bottleneck steps in
the algorithm.  First, computing the normalizing factor in equation
\ref{update} is a bottleneck step. It involves evaluating the product
of two dense matrix $A \cdot B$ at only nonzeros of a sparse matrix
$C$. Earlier we developed a kernel (SDDMM) for this step which is a bottleneck
for several other factor models including our online Variational Bayes LDA
and Sparse Factor Analysis \cite{canny2013big}.

%\subsubsection{Memory Management}
Second, line 7 of the algorithm is another dominant step, and has the
same operation count as the SDDMM step. We wrote a custom kernel that
implements lines 4-11 of the algorithm with almost the same speed as
SDDMM. Finally, we use a matrix-caching strategy \cite{cannybidmach} to
eliminate the need for memory allocation on the GPU in each iteration.

\section{Experiments}
\subsection{LDA}
In this section, we evaluate the performance of the SAME/Factored Gibbs
sampler against several other algorithms and systems. We implemented LDA
using our SAME approach, VB online and VB batch, all using GPU acceleration. 
% Wealso implemented parameter learning for generic Baysian networks with
% cooled Gibbs. 
The code is open source and distributed as part of the BIDMach project
\cite{cannybidmach, canny2013big} on github \footnote{\url{https://github.com/BIDData/BIDMach/wiki}}. We compare our systems
with four other systems: 1) Blei et al's VB batch implementation for LDA
\cite{blei2003latent}, 2) Griffiths et al's collapsed Gibbs sampling (CGS)
for LDA \cite{griffiths2004finding}, 3) Yan et al's GPU parallel CGS for
LDA \cite{yan2009parallel}, 4) Ahmed et al's cluster CGS for LDA \cite{ahmed2012scalable}.

1) and 2) are both C/C++ implementations of
the algorithm \footnote{ 2) provides a Matlab interface for
  CGS LDA} on CPUs. 3) is the state-of-the-art GPU implementation of
parallel CGS. To our best knowledge, 3) is the fastest single-machine LDA
implementation to date, and 4) is the fastest cluster implementation of LDA.

All the systems/algorithms are evaluated on a single PC equipped with
a single 8-core CPU (Intel E5-2660) and a dual-core GPU (Nvidia
GTX-690), except GPU CGS and cluster CGS. Each GPU core comes with 2GB
memory. Only one core of GPU is used in the benchmark. The benchmark
for GPU CGS is reported in \cite{yan2009parallel}. They run the
algorithm on a machine with a GeForce GTX-280 GPU. The benchmarks we
use for cluster CGS were reported on 100 and 1000 Yahoo cluster nodes
\cite{ahmed2012scalable}.

Two datasets are used for evaluation. 1) {\bf NYTimes}. The dataset has approximately 300k New
York Times news articles. There are 100k unique words and 100 million
tokens in the corpus. 2) {\bf PubMed}. The dataset contains about 8.2
million abstracts from collections of US National Library of
Medicine. There are 140k unique tokens, and 730 million words in the
corpus. 

%3) {\bf Dynamic Learning Maps}. The dataset contains a collection (30,000) of students' responses to questions from the DLM Alternate Assessment System. There are 4000 students and 340 questions (97.8\% responses are missing). Each question is derived from a sub-set of 15 concepts. The goal is to estimate the parameters of a given Baysian network representing the relationship between concepts and problems.

%\begin{enumerate}
%
%\item NYTimes . The dataset has approximately 300k New York Times news articles. There are 100k unique words and 100 million tokens in the corpus.
%\item PubMed. The dataset contains about 8.2 million abstracts from collections of US National Library of Medicine. There are 140k unique tokens, and 730 million words in the corpus.
%\end{enumerate}

\subsubsection{Convergence}
We first compare the convergence of SAME Gibbs LDA with other
methods. We choose $m=100$ for the SAME sampler, because convergence
speed only improves marginally beyond $m=100$, however, runtime per
pass over the dataset starts to increase noticeably beyond $m=100$ while being
relatively flat for $m < 100$. The mini-batch size is set to be $1/20$
of the total number of examples. We use the per word log likelihood
$ll=\frac{1}{N^{test}}\log(P(X^{test}))$ as a measure of convergence.

\begin{figure}[t!]
\begin{subfigure}[t!]{0.33\textwidth}
\centering
\includegraphics[scale = 0.28]{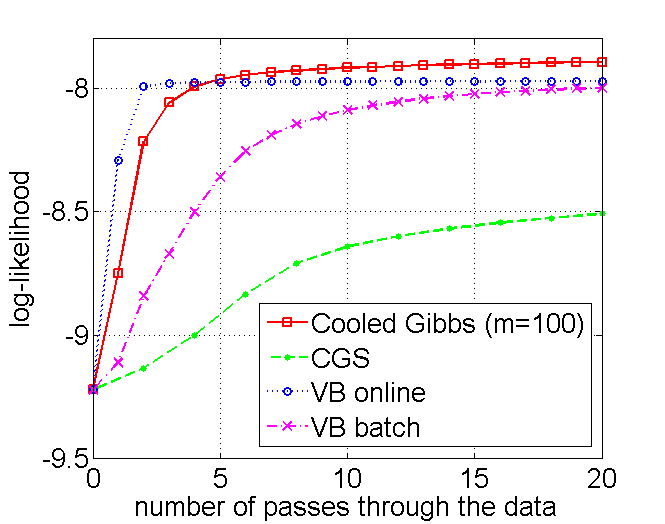}
\caption{ll vs. n passes on NYTimes }
\label{nytimes}
\end{subfigure}  
\begin{subfigure}[t!]{0.33\textwidth}
\centering
\includegraphics[scale = 0.28]{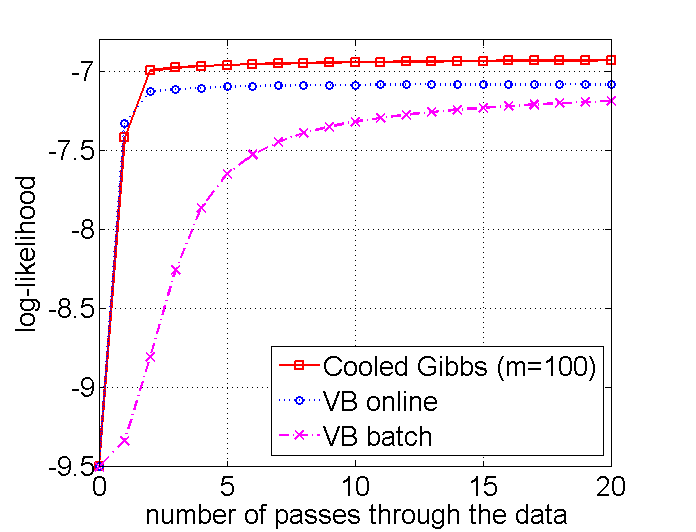}
\caption{ll vs. n passes on PubMed}
\label{pubmed}
\end{subfigure} 
\begin{subfigure}[t!]{0.33\textwidth}
\centering
\includegraphics[scale = 0.28]{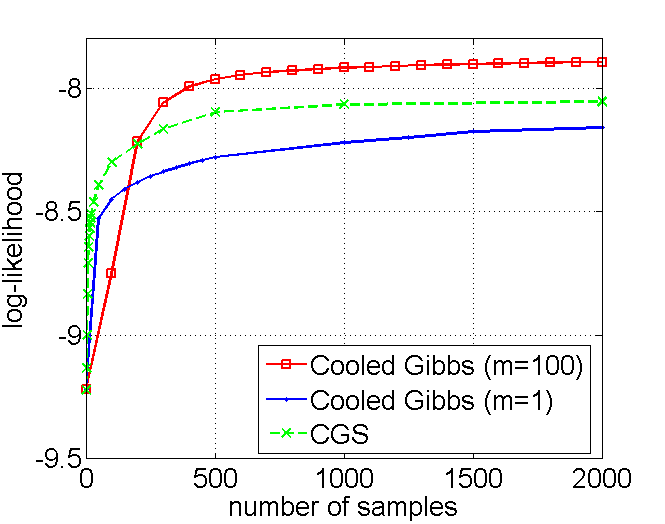}
\caption{ll vs. number of samples/word }
\label{CGS}
\end{subfigure}
\caption{Convergence Comparison} \label{ll}
\label{convergence}
\vspace*{-0.1in}
\end{figure}

Figure \ref{convergence} shows the cross-validated likelihood as a
function of the number of passes through the data for both datasets,
up to 20 passes. As we can see, the SAME Gibbs sampler converges to
a higher quality result than VB online and VB batch on both
datasets. The SAME sampler and VB online converged after 4-5
passes for the NYTimes dataset and 2 passes for the PubMed
dataset. VB batch converges in about 20 passes.
% The algorithms converge faster on the PubMed dataset, because the dataset has more examples for each iteration.

All three methods above are able to produce higher quality than CGS
within 20 passes over the dataset. It usually takes 1000-2000 passes
for CGS to converge for typical datasets such as NYTimes, as can be
seen in Figure \ref{CGS}.

%CGS did not converge in the time we had available (?? hours). We did run 2000 passes of GCS which takes about a week. 

Figure \ref{CGS} plots log-likelihood against the number of samples
taken per word. We compare the SAME sampler with $m=100$, $m=1$ and
CGS. To reach the same number of samples CGS and the SAME sampler
with $m=1$ need to run 100 times as many passes over the data as the
SAME sampler with $m=100$. That is Figure \ref{CGS} shows 20
passes of SAME sampler with $m=100$ and 2000 passes of the
other two methods. At the beginning, CGS converges faster. But in the
long run, SAME Gibbs leads to better convergence. Notice that SAME
with $m=1$ is not identical to CGS in our implementation, because
of minibatching and the moving average estimate for $\Theta$.

%\begin{figure}
%\centering
%\includegraphics[scale = 0.5]{samples.png}
%\caption{standard CGS}
%\label{CGS}
%\end{figure}

\subsubsection{Runtime Performance}

We further measure the runtime performance of different methods with
different implementations for LDA. Again we fix the sample size for
SAME GS at $m=100$. All the runtime number are generated by running each
method/system on the NYTimes dataset with $K=256$, except that Yan's
parallel CGS \cite{yan2009parallel} reports their benchmark for
$K=128$. We also report time to convergence, which is defined as the
time to reach the log-likelihood for standard CGS at the $1000^{th}$
iteration.

\begin{table}
\caption{Runtime Comparison on NYTimes}
\small
\centering
\begin{tabular}[l]{|c|c|c|c|c|c|c|} 
\hline
& \shortstack{SAME GS \\ BIDMach} & \shortstack{VB online \\BIDMach} & \shortstack{ VB batch \\ BIDMach } & \shortstack{VB batch \\ Blei et al } & \shortstack{CGS \\ Griffiths et al } & \shortstack{GPU CGS \\ Yan et al} \\\hline
Runtime per pass (s) & 30 & 20 & 20 & 12600 & 225 & 5.4\\ \hline
Time to converge (s) & 90 & 40 & 400 & 252000 & 225000 & 5400\\ \hline
\end{tabular}
\label{runtime}
\vspace*{-0.15in}
\end{table}

Results are illustrated in the Table \ref{runtime}. The SAME Gibbs
sampler takes 90 seconds to converge on the NYTimes dataset. As
comparison, the other CPU implementations take around 60-70 hours to
converge, and Yan's GPU implementation takes 5400 second to converge
for $K=128$. Our system demonstrates two orders of magnitude
improvement over the state of the art.
% The cooled sampler takes 50 seconds to converge on the PubMed dataset. This is also much faster than any other system to our knowledge. 
Our implementation of online VB takes about 40 seconds to
converge on the NYTimes dataset. The Gibbs sampling method is close in
performance (less than a factor of 3) to that.

Finally, we compare our system with the state-of-art cluster
implementation of CGS LDA \cite{ahmed2012scalable}, using time to
convergence.  Ahmed et al. \cite{ahmed2012scalable} processed 200 million news
articles on 100 machines and 1000 iterations in 2mins/iteration = 2000
minutes overall = 120k seconds. We constructed a repeating stream
of news articles (as did \cite{ahmed2012scalable}) and ran for two 
iterations - having found that this was sufficient for news datasets
of comparable size. This took 30k seconds, which is 4x faster, on
a single GPU node. Ahmed et al. also processed 2 billion
articles on 1000 machines to convergence in 240k seconds, which is 
slightly less than linear scaling. Our system (which is sequential 
on minibatches) simply scales linearly to 300k seconds on this problem. 
Thus single-machine performance of GPU-accelerated SAME GS is almost
as fast as a custom cluster CGS on 1000 nodes, and 4x faster than 100 nodes.

%% Finally, we evaluated the cooled sampler for running parameter
%% estimation on the Dynamic learning dataset. Preliminary results show
%% that the cooled sampler running on CPU is 500x faster than Murphy's
%% bayesnet toolbox. However, 100x is due to our parallel implementation
%% and matrix formulation of the problem. Cooled Gibbs with linear
%% annealing scheduling $m_t = \frac{2t}{T+1}$ ($T$ is total number of
%% passes) is able to achieve 5x faster than standard Gibbs sampling.

\subsubsection{Multi-sampling and Annealing}
\begin{figure}[t!]
\begin{subfigure}[t!]{0.5\textwidth}
\centering
\includegraphics[scale = 0.4]{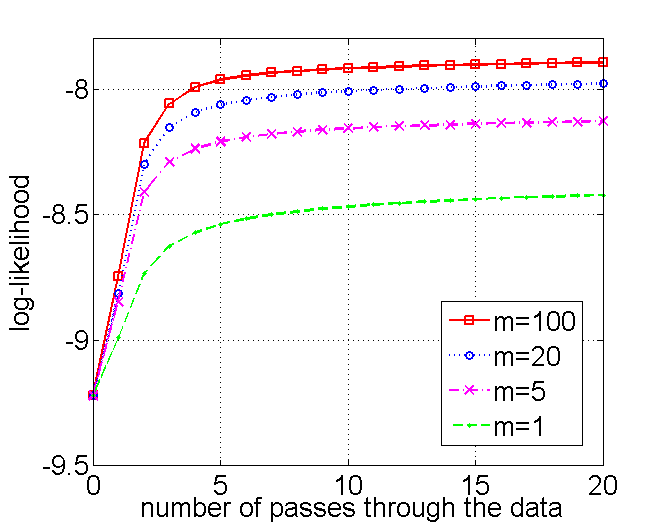}
\caption{ll vs. sample size m}
\label{nsamples}
\end{subfigure}  
\begin{subfigure}[t!]{0.5\textwidth}
\centering
\includegraphics[scale = 0.4]{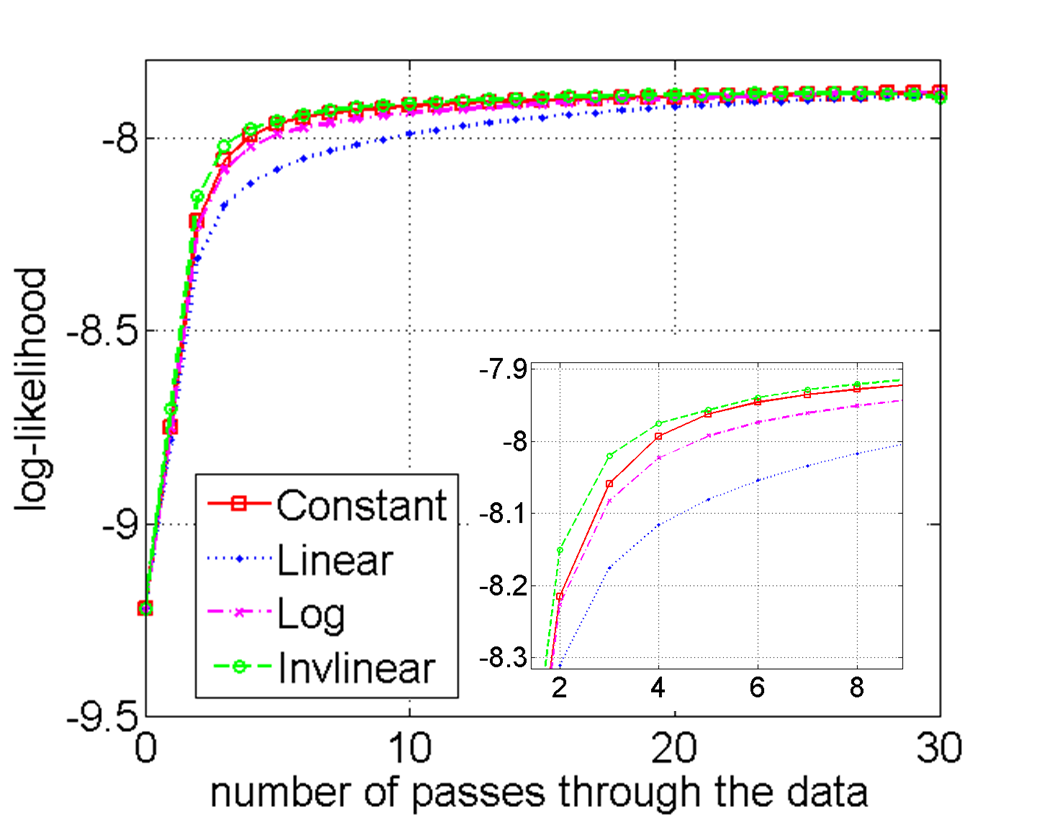}
\caption{annealing schedule}
\label{annealing}
\end{subfigure} 
% \begin{subfigure}[t!]{0.33\textwidth}
% \centering
% \includegraphics[scale = 0.29]{entropy.png}
% \caption{cooling effect}
% \label{variance}
% \end{subfigure}
\caption{Effect of sample sizes} 
\label{samplesize}
\vspace*{-0.1in}
\end{figure}

The effect of sample number $m$ is studied in Figure \ref{samplesize}
and Table \ref{nsample}. As expected, more samples yields better
convergences given fixed number of passes through the data.  And the
benefit comes almost free thanks to the SIMD parallelism offered by
the GPU. As shown in Table \ref{nsample}, $m=100$ is only modestly slower
than $m=1$. However, the runtime for $m=500$ is much longer than
$m=100$.

We next studied the effects of dynamic adjustment of $m$, i.e.
annealing. We compare constant scheduling (fixed $m$) with 
\begin{enumerate}
\itemsep-6pt
\item linear
scheduling, $m_t = \frac{2mt}{t_{max}+1}$, 
\item logarithmic scheduling, $m_t
= m t_{max} \log t /\sum\limits_{t=1}^{t_{max}}\log t$, 
\item invlinear scheduling,
$m_t = \frac{2m(t_{max}+1-t)}{t_{max}+1}$.
\end{enumerate}

 $t_{max}$ is the total number of
iterations. The average sample size per iteration is fixed to $m=100$
for all 4 configurations. As shown in Figure \ref{annealing}, we
cannot identify any particular annealing schedule that is
significantly better then fixed sample size. Invlinear is slighter
faster at the beginning but has the highest initial sample number.

% Finally, we study the cooled distribution we sample from. It is hard to find the distributions for all the parameter configurations. Instead we use the empirical word-topic distributions (on NYTimes dataset) as a proxy. We compute the entropy of the distributions for different words, and different sample size $m$. The entropies are normalized by the entropy at $m=100$ for word ``web''. Lower entropy indicates that the empirical sampling distribution is more concentrated and ``cooled''. Figure \ref{variance} shows the cooling effect. As we can see, the empirical entropies decreases with increasing $m$. The figure provides evidence that the distribution we sample from is cooled for larger $m$. However, sample size $m$ beyond 100 offers only marginal contribution on reducing variance. It also shows that word ``web'' has more spread-out topic distribution of ``school'' and ``book''.

%\begin{figure}
%\centering
%\includegraphics[scale = 0.5]{nsample.png}
%\caption{nsamples}
%\label{nsamples}
%\end{figure} 

\begin{table}
\caption{Runtime per iteration with different $m$}
\small
\centering
\begin{tabular}[l]{|c|c|c|c|c|c|} 
\hline
& m=500 & m=100 & m=20 & m=5 & m=1\\\hline
Runtime per iteration (s) & 50 & 30 & 25 & 23 & 20  \\ \hline
\end{tabular}
\label{nsample}
\vspace*{-0.2in}
\end{table}

\section{Conclusions and Future Work}
This paper described hardware-accelerated SAME Gibbs Parameter
estimation - a method to improve and accelerate parameter estimation
via Gibbs sampling. This approach reduces the number of passes over
the dataset while introducing more parallelism into the sampling
process. We showed that the approach meshes very well with SIMD
hardware, and that a GPU-accelerated implementation of cooled GS for
LDA is faster than other sequential systems, and comparable with the
fastest cluster implementation of CGS on 1000 nodes. The code is at
\verb+https://github.com/BIDData/BIDMach+

SAME GS is applicable to many other problems, and we are currently
exploring the method for inference on general graphical models.  The
coordinate-factored sampling approximation is also being applied to
this problem in conjunction with full sampling (the approximation
reduces the number of full samples required) and we anticipate
similar large improvements in speed. The method provides the
quality advantages of sampling and annealed estimation, while
delivering performance with custom (symbolic) inference methods. 
We believe it will be the method of choice for many inference
problems in future. 

\eject

{\small
\bibliographystyle{abbrv}
\bibliography{refs}
}
\end{document}